\documentclass{article}

\usepackage{rotating}
\usepackage{longtable}
\usepackage{booktabs}
\usepackage{multirow}
\usepackage[load-configurations=version-1]{siunitx}
\usepackage{graphicx}
\usepackage{mathtools}
\usepackage{algorithm}
\usepackage[noend]{algorithmic}
\usepackage{color}
\usepackage{url}

\usepackage{array}
\usepackage{amsmath}
\usepackage{microtype}
\usepackage{pdflscape}

\begin{document}

\title{On the Use of Default Parameter Settings in the Empirical Evaluation of Classification Algorithms}

\author{Anthony Bagnall and Gavin C. Cawley \\
School of Computing Sciences \\
              University of East Anglia    \\
              Norwich NR4 7TJ              \\
              United Kingdom
              }

\maketitle

\begin{abstract}
We demonstrate that, for a range of state-of-the-art machine learning algorithms, the differences in generalisation performance obtained using default parameter settings and using parameters tuned via cross-validation can be similar in magnitude to the differences in performance observed between state-of-the-art and uncompetitive learning systems.  This means that fair and rigorous evaluation of new learning algorithms requires performance comparison against benchmark methods with best-practice model selection procedures, rather than using default parameter settings.  We investigate the sensitivity of three key machine learning algorithms (support vector machine, random forest and rotation forest) to their default parameter settings, and provide guidance on determining sensible default parameter values for implementations of these algorithms. We also conduct an experimental comparison of these three algorithms on 121 classification problems and find that, perhaps surprisingly, rotation forest is significantly more accurate on average than both random forest and a support vector machine.
\end{abstract}


\section{Introduction}

\label{sec:intro}
Dr Bunsen Honeydew, of Muppet Labs, recounts an anecdote in which he had developed a novel
binary pattern recognition method, namely the Muppet Labs Machine Learning Algorithm
([ML]$^2$A).  To demonstrate the competitiveness of this approach he performed an extensive
empirical  evaluation over a suite of benchmark datasets with  multiple randomised
partitioning to form the training and test sets.   The performance the of the [ML]$^2$A was
compared with that of a range of state-of-the-art machine learning algorithms, namely the
Support Vector Machine (SVM) \cite{Boser1992a,Cortes1995a} and Least-Squares Support Vector
Machine (LS-SVM) \cite{Suykens2002a}, both using the spherical Radial Basis Function (RBF)
kernel and the Expectation-Propagation based Gaussian Process Classifier (EP-GPC)
\cite{Rasmussen2006a}, with the isotropic squared exponential covariance function.  Following
the recommendation of Dem\v{s}ar \cite{Demsar2006a}, he used the Friedmann test to determine
if there were any statistically significant differences in the rankings of the classifiers.
However, following recent recommendations in \cite{benavoli16posthoc} and
\cite{garcia08pairwise}, he abandoned the Nemenyi post-hoc test originally used by
\cite{Demsar2006a} to form cliques (groups of classifiers within which there is no
significant difference in ranks). Instead, he compared all classifiers with pairwise Wilcoxon
signed rank tests, and formed cliques using the Holm correction (which adjusts family-wise
error less conservatively than a Bonferonni adjustment).

\begin{figure}[bh!]
   \begin{center}
      \includegraphics[width=10cm, trim={4cm 14cm 4cm 10cm}]{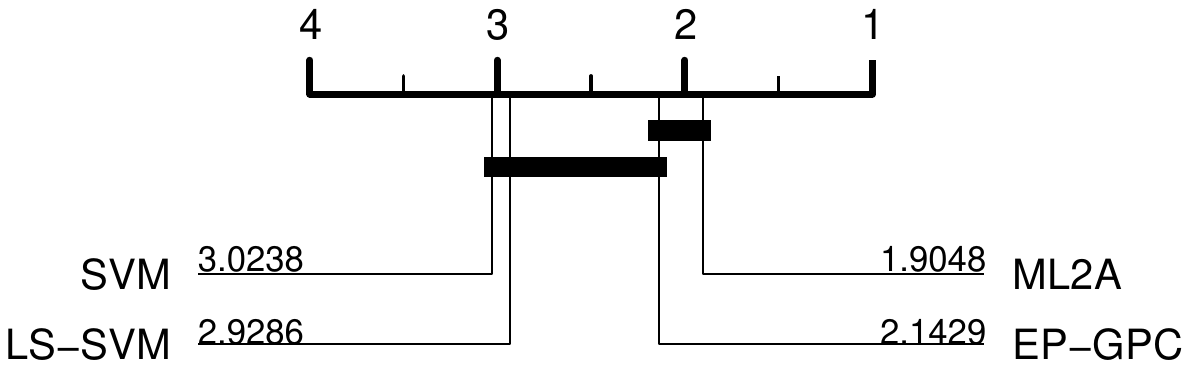}
   \end{center}
   \caption{Critical difference diagram \cite{Demsar2006a}, showing the mean
            ranks of four classifiers, over a suite of benchmark datasets, obtained by Dr Bunsen Honeydew. The solid bars group classifiers into cliques, within which there is no pairwise statistical difference in ranks. }
   \label{fig:Bunsen_results}
\end{figure}

Based on the results, summarised in Figure~\ref{fig:Bunsen_results}, [ML]$^2$A
appeared highly promising; the [ML]$^2$A achieves the highest overall ranking, was found to
be competitive with EP-GPC and statistically superior to both the SVM and LS-SVM.  Dr
Honeydew swiftly began drafting a paper for a prestigious journal...

\begin{figure}[th!]
   \begin{center}
      \includegraphics[width=10cm, trim={4cm 14cm 4cm 10cm}]{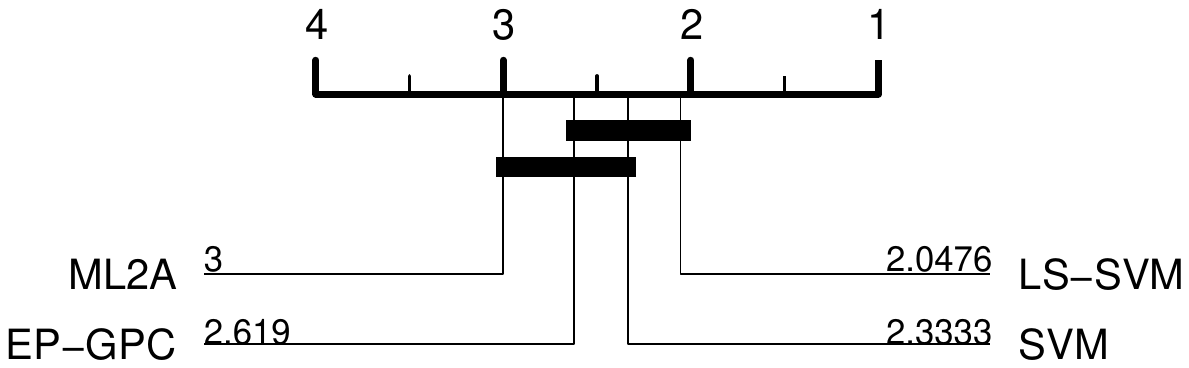}
   \end{center}
   \caption{Critical difference diagram \cite{Demsar2006a}, showing the mean
            ranks of four classifiers, over the suite of benchmark datasets
            shown in Table~\ref{tbl:datasets}, obtained by Dr Honeydew's
            research assistant Beaker.}
   \label{fig:Beaker_results}
\end{figure}

However, being both diligent and cautious, Dr Honeydew first asked his research assistant,
Beaker, to replicate his results, just to be sure.  Beaker reimplemented the [ML]$^2$A from
scratch and both were reassured to find that it gave exactly the same results on the
benchmark datasets.  However, Beaker obtained a very different set of results for the overall
comparison, shown in Figure~\ref{fig:Beaker_results}.  Beaker found that the [ML]$^2$A
achieved the \emph{lowest} overall ranking, and while it was competitive with the EP-GPC and
SVM, it was statistically inferior to the LS-SVM.  As a result, Dr Honeydew was reluctantly
forced to reconsider his publication plan!

So, how could such different sets of results be obtained from such thorough empirical
evaluations?  It transpires that the difference lay in the way in which the values of the
hyper-parameters of the benchmark classifiers, i.e. the kernel/covariance function and
regularisation parameters, were determined.  Performing an experimental evaluation using
multiple classifiers, multiple benchmark datasets and multiple randomised partitions is
computationally expensive.  Dr Honeydew therefore decided to use the same default parameter
settings for the SVM, LS-SVM and EP-GPC for all benchmark datasets; all kernel/covariance
function and regularisation parameters were set to one.  Beaker on the other hand, being more
adept at parallel programming and the use of High Performance Computing (HPC)
facilities, tuned the hyper-parameters for each method independently, for each test/training
partition of each benchmark dataset to minimise a cross-validation based model selection
criterion.

While the setting of this example is fictional, the experimental results are real, and full
details are given in Appendix~\ref{sec:muppet}.  In reality, the [ML]$^2$A is actually a
simple multi-layer perceptron neural network with Bayesian regularisation
\cite{Bishop1995a,Nabney2004b}.  Naturally, Beaker's evaluation protocol provides the more
reliable indication of the relative performance of the classifiers, simple MLP classifiers
are unlikely to outperform more modern kernel learning methods on average.
This demonstrates that the use of default parameters in experimental evaluation of machine
learning algorithms is unsatisfactory and the practice should be deprecated.


We have repeatedly seen this type of bias in machine learning research.  A large proportion of machine learning research involves proposing alternative classification algorithms or novel variants of existing algorithms. The new algorithms are usually compared to existing algorithms through an experimental evaluation on some subset of the machine learning repository hosted by University of California, Irvine. One of the prime criteria for algorithm assessment is classification accuracy (or error). There are now several fairly mature software suites available to facilitate a sound comparison such as WEKA and R packages. These allow a researcher to build classifiers without the need for an in depth understanding of how the classifier works. This has massively widened the user base for machine learning advances. However, there is always a danger of using algorithms without a proper understanding of how they actually work, namely that the various classifier systems may be applied with differing levels of skill (perhaps not even competently) which biases any performance evaluation in favour of the learning systems with which the user is most familiar. It is unfortunately common to compare  established classification algorithms using the default parameters provided in the implementation. But what if the default parameters are poor, or the algorithm is particularly sensitive to parameter values? The conclusions drawn about the supposed algorithmic advance are likely to be unreliable.

The remainder of the paper is structured as follows:  Section~\ref{wekaSVM} investigates the sensitivity of the Support Vector Machine with the usual RBF kernel function to the setting of the hyper-parameters, finding that the default values used in the popular WEKA package are far from optimal, such that it could only provide a straw man baseline in a performance comparison.  Section~\ref{wekaForest} goes on to demonstrate that the Random Forest classifier is also sensitive to the default parameters, although to a lesser extent than observed in the case of the RBF SVM.  A ``bakeoff'' comparing the performance of state-of-the-art and uncompetitive classifier systems is given in Section~\ref{bakeoff}, demonstrating the importance of parameter tuning in performance evaluation.  These experiments also demonstrated the unexpectedly good performance of the Rotation Forest method, which appears to have been previously underestimated, perhaps due to poor default parameter settings.  Section~\ref{better_defaults} discusses the potential for better default parameter settings.  Finally, the work is summarised and conclusions drawn in Section~\ref{conclusions}.

\section{Support Vector Machines in WEKA}
\label{wekaSVM}
We can demonstrate the problems with default parameters with the WEKA~\cite{hall2009weka} implementation of SVM, which uses the sequential minimal optimization  algorithm~\cite{platt98sequential}. It converts nominal attributes to binary ones, normalises all attributes by default, and uses pairwise classification for multi-class problems. It can be used with a range of kernels, but it defaults to a linear kernel with margin parameter, $C$, set to 1. The classifier class is called \texttt{SMO}.

Suppose we wish to test the hypothesis that on average an RBF kernel is more accurate than a linear kernel. To test this hypothesis, we use the suite of 121 UCI data sets used to create the results presented in~\cite{delgado14hundreds} and available online\footnote{the data is available from http://persoal.citius.usc.es/manuel.fernandez.delgado/papers/jmlr/data.tar.gz}. A single run involves performing 30 random train/test resamples on a single data set and measuring the average accuracy over these resamples.  We repeat this over all data, then perform both parameteric and non-parametric statistical tests on the results of different classifiers.

\begin{figure}[!ht]
	\centering
\begin{tabular}{cc}
       \includegraphics[width=6cm, trim={4cm 10.5cm 0cm 0cm}]{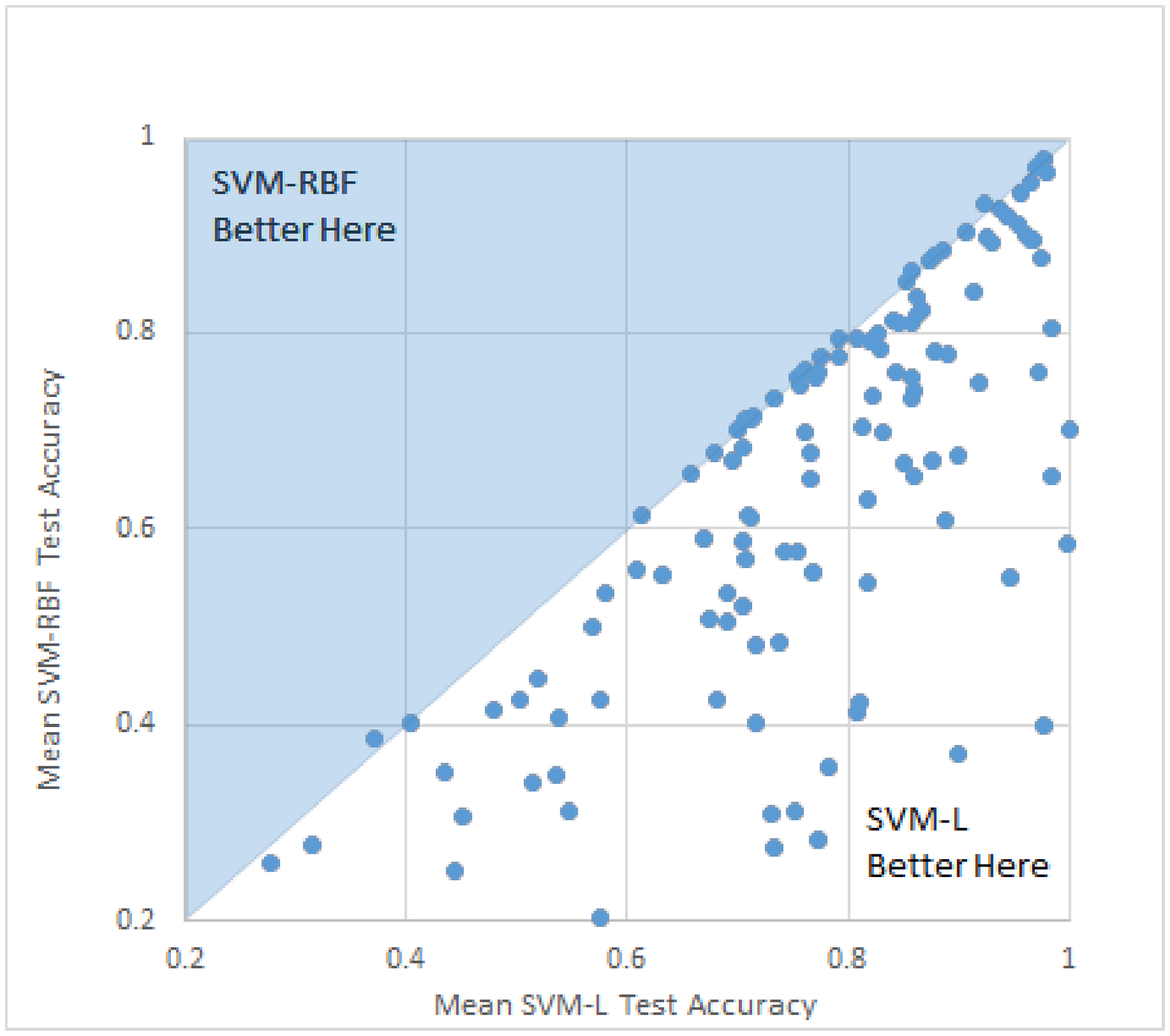}
       &
       \includegraphics[width=6cm,trim={4cm 10.5cm 0cm 0cm}]{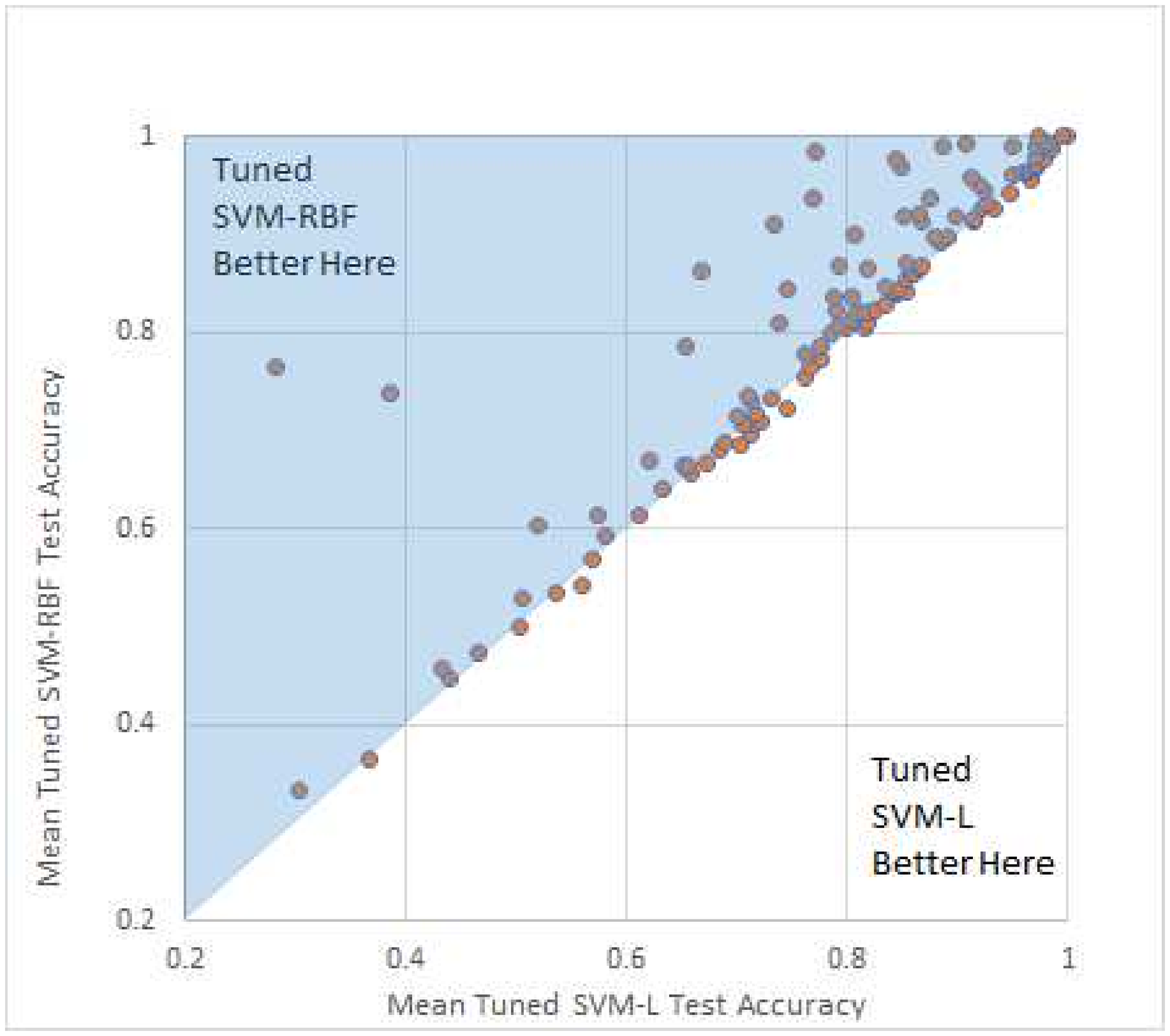}\\
       (a) & (b) \\
\end{tabular}
       	\caption{A scatter plot of accuracies of WEKA's SMO classifier with a RBF kernel and a linear kernel with default parameters. Figure (a) compares untuned classifiers, figure (b) plots the accuracy of tuned classifiers.}
       \label{fig:smv1}
\end{figure}

For the first experiment we use the default parameters for the linear and RBF kernel with the WEKA SMO classifier. Received wisdom is that an RBF kernel will give a more accurate classifier on average, and this has been observed in previous experimental studies. Note an SVM with a linear kernel can be approximated by an SVM with the RBF kernel with appropriate values for the hyper-parameters \cite{keerthi2003asymptotic}, and so in principle should perform at least as well.  A recent experimental study assessed 179 classifiers on 121 data sets~\cite{delgado14hundreds}. The main conclusion of this work was that {\em``The classifiers most likely to be the bests {[sic]} are the random forest (RF) versions"} but that {\em ``the difference is not statistically significant with the second best, the SVM with Gaussian kernel"}. Our prior belief then is that at worst there would be no significant difference between a SVM with linear and RBF kernels. The results, displayed in Figure~\ref{fig:smv1}(a), completely contradict our prior beliefs. The linear SVM wins on 101 problems, ties on 8 and loses on just 12. The mean difference in accuracy is over 12\%, the median difference over 7.5\%. The difference is statistically significant with any test you care to mention.

So why is WEKA's SVM with the RBF kernel so bad?  It is unlikely that there is a bug in the code; WEKA is heavily used code, the RBF kernel is not hard to implement and there is nothing wrong in the code that we can see. To put this into context, a 1-nearest neigbour (1-NN) classifier is significantly more accurate than WEKA's SMO with default RBF kernel: 1-NN beats RBF on 82 out of 121 problems with a median difference of almost 7\% (the linear SVM is not significantly different to 1-NN).

\begin{figure}[!ht]
	\centering
       \includegraphics[width=10cm, trim={4cm 14cm 4cm 12cm}]{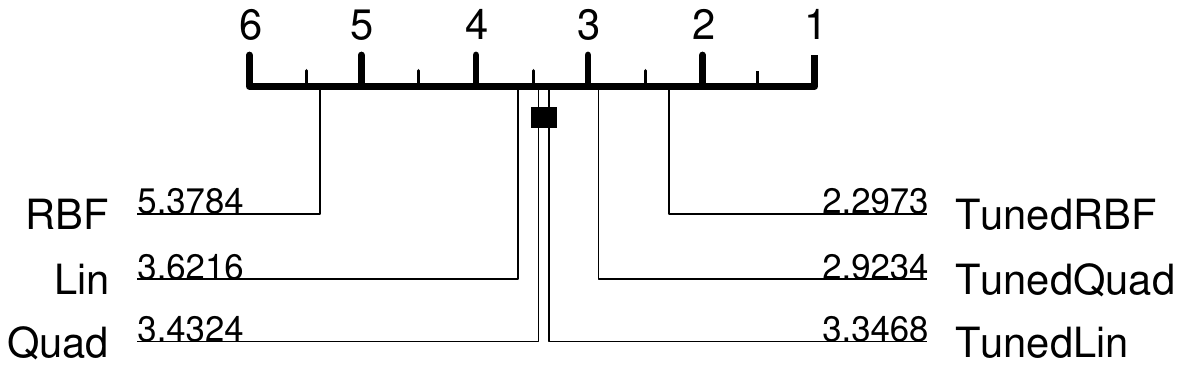}
       	\caption{Critical difference diagram for WEKA's SMO classifier with linear, quadratic and RBF kernel, both tuned and untuned.}
       \label{fig:smv2}
\end{figure}

We would expect a significant improvement in performance if we tune the parameters for both kernels on the train data. We perform a ten-fold cross-validation for the parameters $C \in \{2^{-16},2^{-15}, \ldots, 2^{8}\}$ for the linear kernel and all pairs $(C,\gamma) \in \{(2^{-16},2^{-16}), (2^{-16},2^{-15}), \ldots, (2^{8},2^{8}) \}$. This is biased towards RBF, because RBF gets 625 tuning evaluations on each fold, whereas linear gets just 25, however this could also potentially introduce some over-fitting of the model selection criterion \cite{Cawley2010b}.  We could adjust for this bias by allowing more evaluations for the linear kernel (or, more practically, reducing the number for RBF), but we are not attempting to describe the best way of tuning parameters. Our goal is merely to demonstrate the huge effect parameter tuning can have on classifier performance. Figure~\ref{fig:smv1}(b) shows how tuning completely reverses the relative performance of linear and RBF SVM. RBF now wins on 77 problems and is on average 2.8\% more accurate. For completeness, We also include a default and tuned quadratic kernel in our analysis.  Figure~\ref{fig:smv2} shows the critical difference diagram.

\begin{figure}[!h]
	\centering
    \begin{tabular}{cc}
       \includegraphics[width=6cm, trim={4cm 2cm 1cm 2.5cm}]{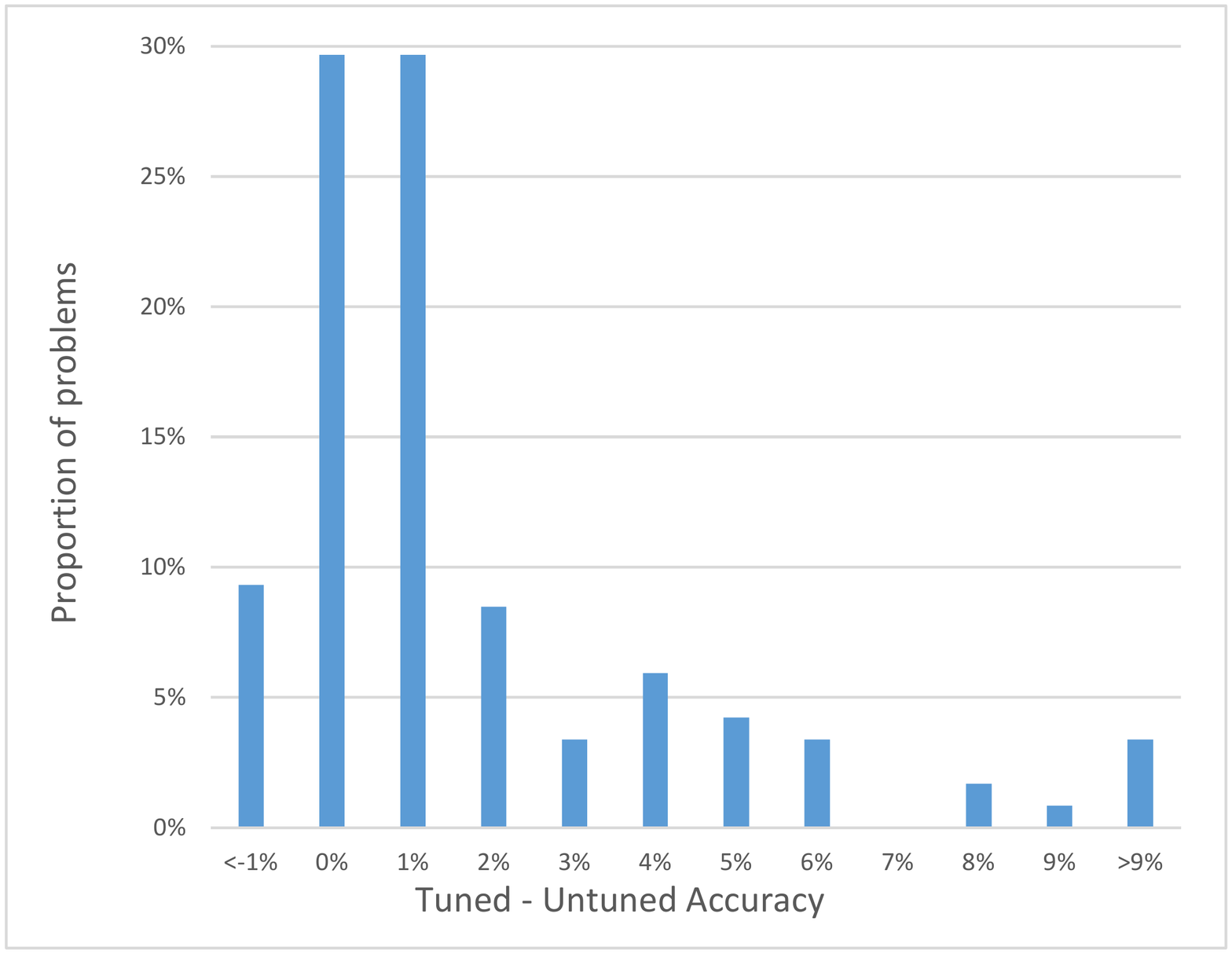} &
       \includegraphics[width=6cm, trim={4cm 2cm 1cm 2.5cm}]{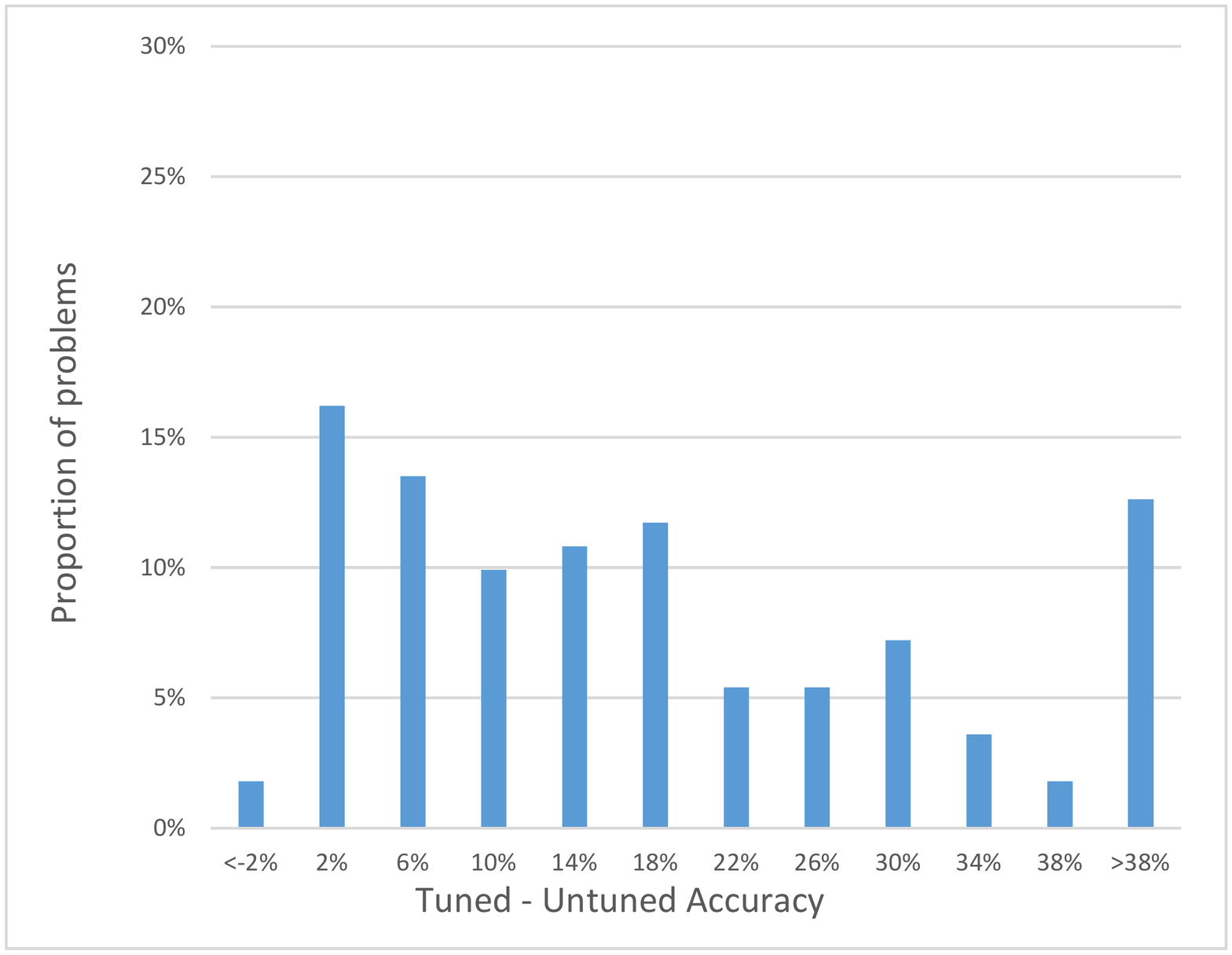}\\
       (a) & (b)
	\end{tabular}
\caption{Histograms of difference in mean accuracy between tuned and untuned SVM classifiers with (a) linear kernel and (b) RBF kernel.}
       \label{fig:smv3}
\end{figure}

There is clearly a significant improvement for all three kernels, but Figure~\ref{fig:smv2} does not demonstrate exactly how big that improvement is. Figure~\ref{fig:smv3} summarises the improvement over all data sets for both linear and RBF kernel by showing the distribution of the difference between tuned and untuned SVM. Tuning improves the linear SVM on 62\% of problems with a mean improvement of just 1.8\% accuracy. Contrast this modest improvement with RBF, where tuning improved accuracy on over 90\% of problems with a remarkable mean difference of 16.5\%. The most ludicrous example is statlog-vehicle (a four class problem), where the accuracy improves from 30.91\% to 98.99\%.
\footnote{The results are available in the spreadsheet www.timeseriesclassification.com/svmCompare.xls.}

If we were using a newly developed in-house implementation of the SVM, results like these would be suggestive of a bug in the code.  However WEKA is widely used and so can be expected to be highly reliable as a bug this severe would almost certainly have been discovered by the existing user-base. Bug or not, it is clear that it is inappropriate to use WEKA's SMO with RBF as a baseline classifier unless the parameters are tuned. There are of course tools for parameter search for WEKA (for example, \cite{kotthoff16autoweka,kuhn08caret} provide a range of tools for WEKA and R respectively) but these are not shipped in the standard implementation. For illustrative purposes we have implemented a classifier {\texttt{TunedSMO}} that extends the \texttt{SMO} classifier to have the option to tune the parameters. This implementation does a simple grid search and is in no way optimised for minimizing memory or time complexity, so should be used with caution (it is probably more sensible to use the built in parameter search routines provided in WEKA~\cite{kohavi95wrappers}). The code to recreate the exact experiments presented in this section is available in the class  \texttt{SVMExperiments} (see the \texttt{generateAll()} method for guidance) available from the accompanying website\footnote{www.timeseriesclassification.com/defaultParas.php}.

\section{Forests in WEKA}
\label{wekaForest}

A key finding of the experimental study conducted by Delgado {\em et al.}~\cite{delgado14hundreds} is that {\em ``the classifiers most likely to be the best are the random forest"}. Leaving aside the validity of this claim for the moment (see~\cite{Wainberg16randomForests} for a discussion of these results), an obvious anomaly with their results is the difference in performance between random forest implementations.
Seven random forests implementations are evaluated. Five are tuned using the Claret interface to R, but the basic R implementation and the WEKA implementation are seemingly very similar. The two main parameters in random forest are the number of trees ($nTrees$) and the number of features to consider in random feature selection for each tree ($nFeatures$). The R version used in~\cite{delgado14hundreds} is the R function randomForest with  $nTrees=500$ and $nFeatures=\sqrt(numAttributes)$. The WEKA version used is stated as having $nTrees=500$ and  $nFeatures= \log_2(numAttributes+ 1)$.
The R version is ranked 5th overall, whereas the WEKA version is 25th, only just beating a linear SVM. The difference in mean accuracy between the R version and the WEKA version is statistically significant: the R version wins on 77 datasets, the WEKA version just 28. Random forest is an elegantly simple algorithm, so it seems highly unlikely that this difference could be attributed to fundamental differences in implementation. The only apparent difference is in the parameter $nFeatures$.  Our experience and received wisdom~\cite{oshiro12nowmany} tells us random forest is robust to this parameter, hence we would not expect the difference in $nFeatures$ to create such a massive difference in performance. To test this, we ran WEKA random forest (500 trees) and $nFeatures=\sqrt(numAttributes)$ and $nFeatures= \log(numAttributes+ 1)$, and found that the classifier  was significantly better using $nFeatures=\sqrt(numAttributes)$.\footnote{see timeseriesclassification.com/RandF500.xls for details}. This seemingly minor difference in default parameters causes a significant difference in accuracy. Our WEKA random forest results are not significantly different to the R random forest results reported by Delgado {\em et al.}.

\begin{figure}[!ht]
	\centering
    \begin{tabular}{cc}
       \includegraphics[height=6cm, angle=90, trim={8cm 2cm 8cm 2cm}]{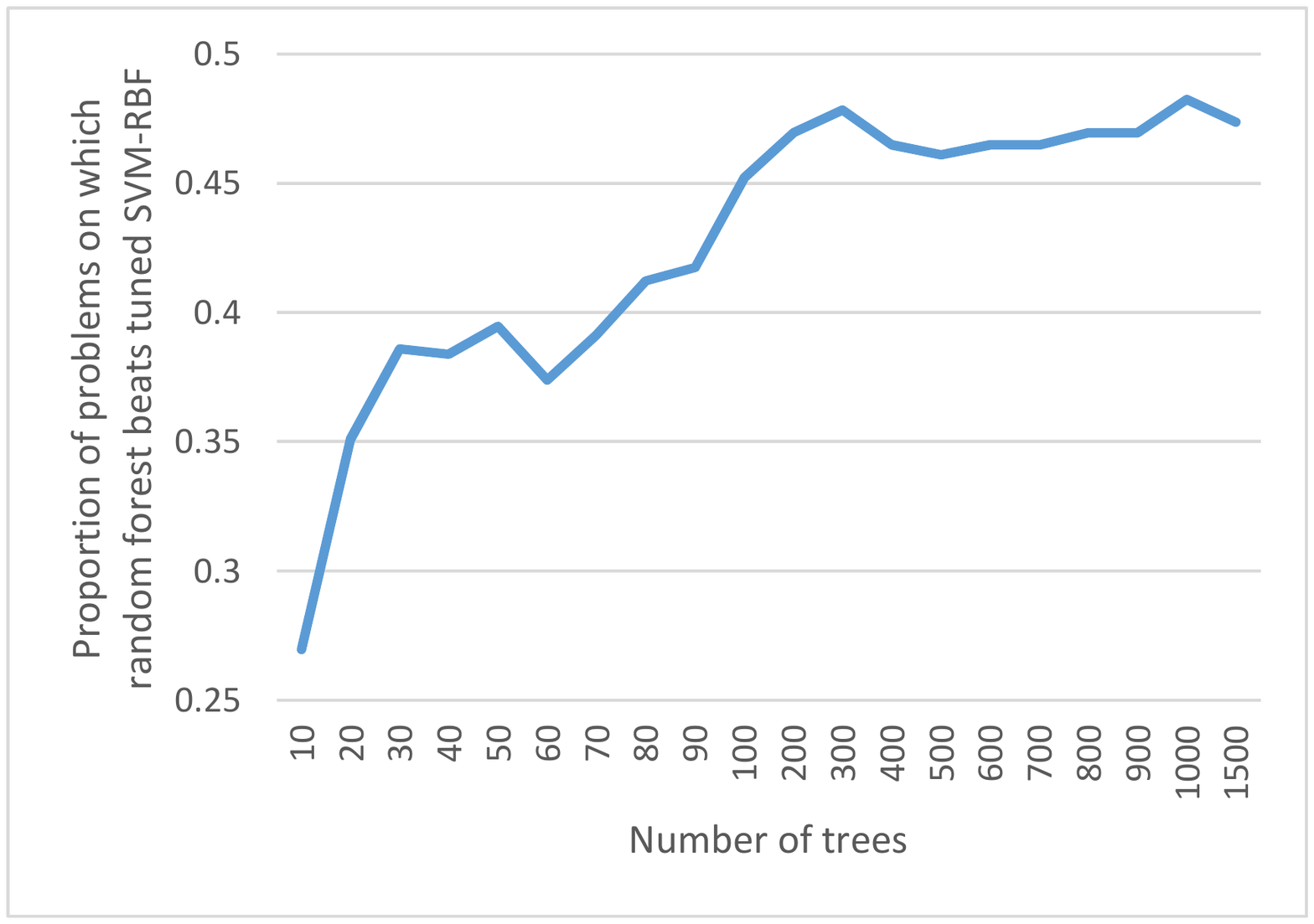} &
       \includegraphics[width=6cm, trim={2cm 8cm 2cm 8cm}]{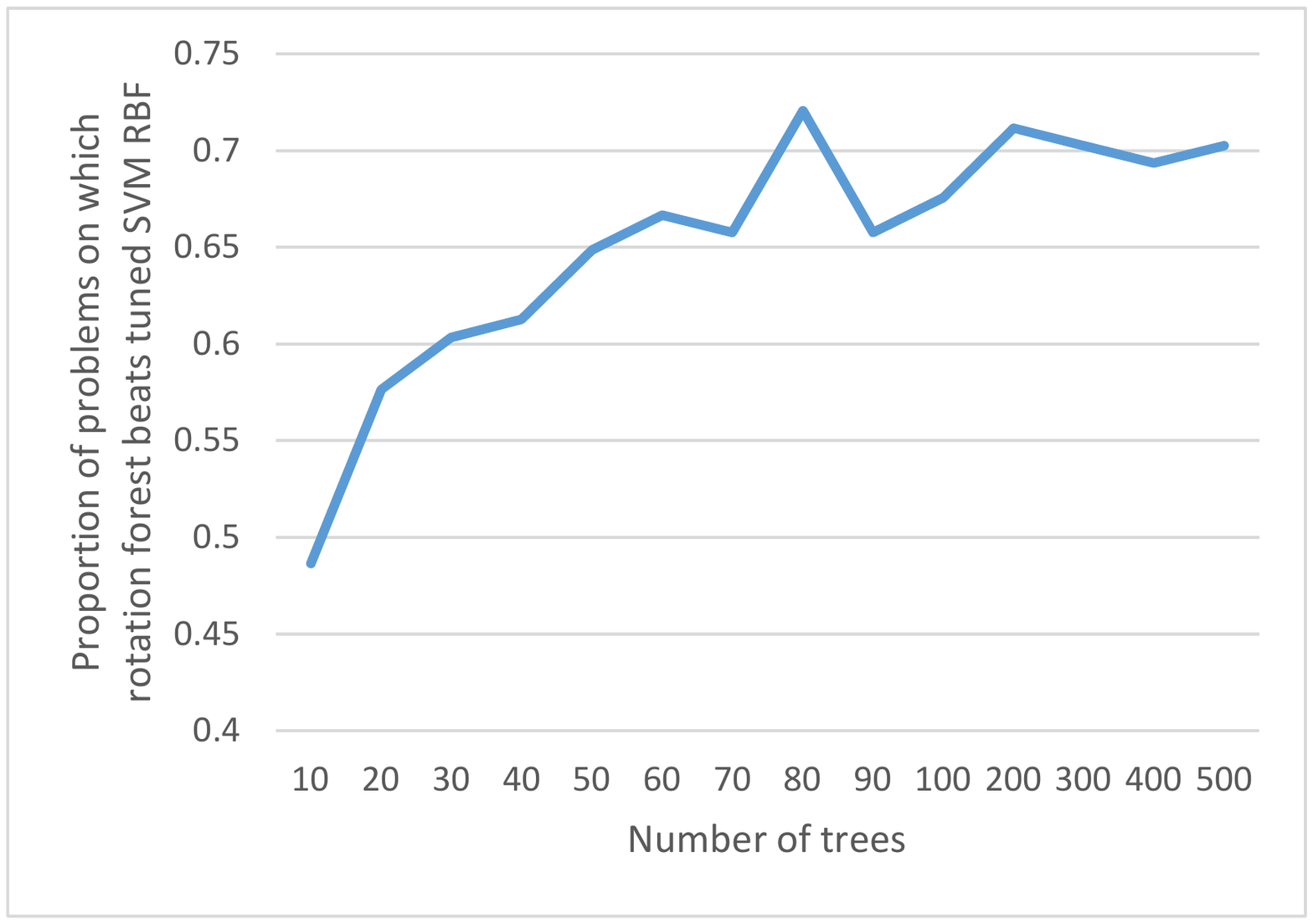} \\
(a) & (b)
\end{tabular}
\caption{Comparison of tuned SVM-RBF with (a) random forest and (b) rotation forest with a varying number of trees.}
       \label{fig:RandF_trees}
\end{figure}

To investigate this further, we run the WEKA random forest  with $ nTrees \in \{10,50,100,200,300,\ldots,1000,1500\}$ and $nFeatures=\sqrt(numAttributes)$. We compare the results for the random forest to those obtained for the tuned SVM-RBF in Section~\ref{wekaSVM}. Figure~\ref{fig:RandF_trees}(a) shows the relative number of problems where random forest beats SVM with a tuned RBF kernel, excluding ties. At the WEKA default value of 10 trees, random forest is significantly worse than SVM, winning on just 31 problems and is on average over 2\% less accurate. This difference reduces until random forest has 200 trees, where there is no significant difference between SVM-RBF and random forest.

The default WEKA parameters give a particularly distorted view of the SVM-RBF and Random Forest classifiers. Another classifier with deceptive results with default parameters is rotation forest~\cite{rodriguez06rotation}.  In our experience with other classification problems~\cite{lines16hivecote,bagnall16bakeoff,bagnall15cote}, rotation forest has always performed better than, or at least not significantly worse than, both SVM and random forest~\cite{lines16hivecote}. However,  in the results presented in~\cite{delgado14hundreds} it only ranks 15th. The default number of trees for rotation forest is 10. Although a very small size for a tree ensemble, it is the default value recommended in the original paper~\cite{rodriguez06rotation}. In~\cite{bagnall16bakeoff} we use random forest with 50 trees, but we have never quantified the sensitivity of the classifier to the number of trees. To do so, we repeat the previous experiment and compare rotation forest to tuned SVM-RBF. The relative performance of rotation forests with ${10,20, \ldots,100,200,\ldots,500}$ base trees are shown in   Figure\ref{fig:RandF_trees}(b).\footnote{see timeseriesclassification.com/Forest.xls for the results of the random forest and rotation forest experiments for this section. To recreate the experiments, see class  \texttt{ForestExperiments} and the \texttt{generateAll()} for guidance.}

For the default value, rotation forest is not significantly different to tuned SVM-RBF. However, with 50 trees or more, it is more accurate than SVM on 65\% of the problems. Once more, we see the default parameters giving a very deceptive impression of a classifier. The rotation forest classifier is seemingly at least competitive, and possibly significantly better, than state-of-the-art classification algorithms, and it has comparable time and space complexity. However, it has received far less attention in the research community than random forest and support vector based algorithms. We believe this is primarily due to the poor default parameters chosen no doubt out of computational expediency.

\section{A Bakeoff of Multiple Untuned and Tuned Classifiers}
\label{bakeoff}

\begin{figure}[!ht]
	\centering
       \includegraphics[width=10cm, trim={4cm 14cm 4cm 12cm}]{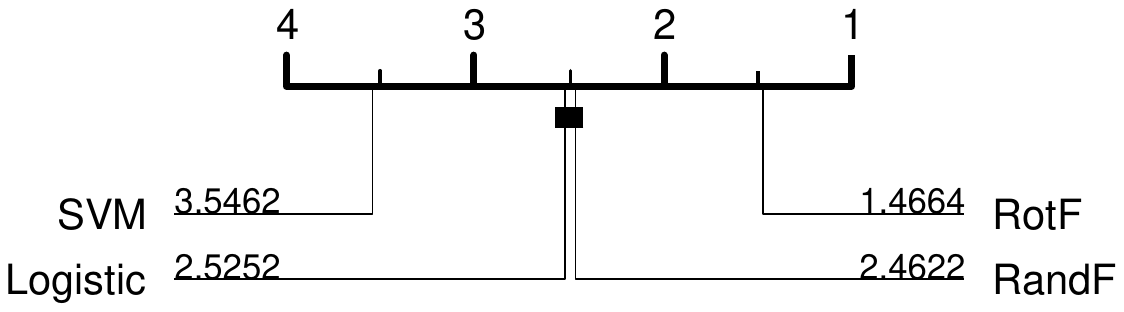}
       	\caption{Critical difference diagram for logistic regression, random forest, rotation forest and SVM-RBF classifiers with no parameter tuning.}
       \label{fig:untuned}
\end{figure}

It may be belabouring the point, but we conduct one further round of experiments to demonstrate the importance of not using default parameters, and as a side effect we more rigorously assess the relative accuracy of tuned SVM, random forest and rotation forest classifiers. Logistic regression is one of the oldest and most widely used classifiers~\cite{cox58logistic}. Given the huge amount of research into classification algorithms in the sixty years since logistic regression was proposed, we would expect contemporary algorithms to be more accurate on average. Figure~\ref{fig:untuned} shows the critical difference diagram for logistic regression and RBF-SVM, random forest and rotation forest with WEKA default parameters. The WEKA version of logistic regression (classifier \texttt{Logistic}) uses multinomial logistic regression model with a ridge estimator, set by default to be very small.

Logistic regression is significantly more accurate than default SVM-RBF, not significantly different to default random forest and significantly worse than rotation forest. The median difference between logistic regression and rotation forest is just over 2\%. We get a similar pattern of results using other basic classifiers such as 1-NN, Naive Bayes or C4.5. Figure~\ref{fig:untuned} seems to support the argument made in~\cite{Hand2006a} that there has been little progress in classifier technology. However, we know that we are being unfair to the contemporary classifiers and that tuning will improve performance. We first need to determine the parameter search method and search space. It is always possible that these  meta-parameters will bias the evaluation, so our principle is to keep it as simple and transparent as possible. We use a grid search for all three classifiers and use an identical evaluation technique;  we evaluated all classifier/parameter setting combinations using a ten fold cross validation on the train data, evaluating on the test data once only. We could have made the random forest much faster by using out of bag error, but that would introduce another degree of freedom into the experiment. Similarly, we could have used a search technique rather than grid search, but then we may end up assessing the regularity of the parameter space rather than the overall ability of the classifier. The parameters we search and their ranges are given in Table~\ref{tab:parameters}.

\begin{table}
\caption{Bakeoff parameter tuning}
\label{tab:parameters}
\begin{tabular}{|ccc|}\hline
Classifier & Parameter & Range \\ \hline
SVM & Kernel		& RBF \\
SVM & Regularisation $C$ & ${2^{-16},2^{-15},\ldots,2^{8}}$ \\
    & Gaussian variance $\gamma$ & ${2^{-16},2^{-15},\ldots,2^{8}}$ \\ \hline
Random Forest & number of trees & ${10,50,100,200,\ldots,1000,1250,1500,1750,2000}$ \\
			& feature subset size & $\sqrt{m}$ \\ \hline
Rotation Forest & number of trees & ${10,50,100,200,\ldots,1000,1250,1500,1750,2000}$\\
			& feature partition size & 3 \\
			& sample proportion & 0.5 \\
\hline
 \end{tabular}
\end{table}

Inevitably, we have made some compromises in this experimental set up. For example: these are not the only parameters we could tune; we have chosen fairly arbitrary ranges and intervals; and we are giving more evaluations to some classifiers than others. All these are valid criticisms, especially the last, but we believe we have covered the parameters and ranges that have the biggest impact on accuracy. We have paid little attention to time and space complexity.
However, it is worth noting that random forest was faster than rotation forest and rotation forest was faster than SVM, which had by far the largest parameter search space (625 combinations instead of 16).

\begin{figure}[!ht]
	\centering
       \includegraphics[width=10cm, trim={4cm 14cm 4cm 12cm}]{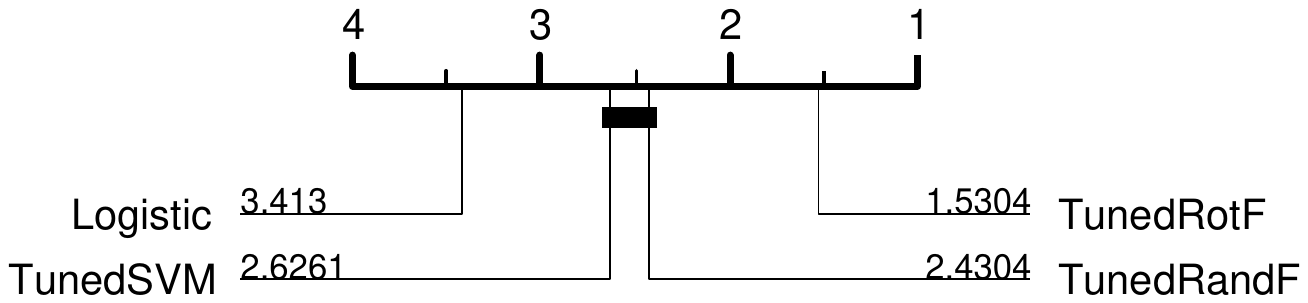}
       \caption{Critical difference diagram for logistic regression, random forest, rotation forest and SVM-RBF classifiers with parameter tuning for the latter three.}
       \label{fig:tuned}
\end{figure}

The results are summarised in the critical difference diagram shown in Figure~\ref{fig:tuned}. Tuned SVM-RBF, random forest and rotation forest are all significantly better than logistic regression. There is no significant difference between random forest and SVM. This result is in agreement with previous experimental studies~\cite{delgado14hundreds}. However, rotation forest is significantly better than both random forest and SVM. We do not believe this result has been previously observed in an experimental study. This omission can, we believe, be attributed to the fact previous studies have used the default parameter value of $nTrees =10$. Rotation forest is significantly better than both SVM and random forest when tested with a paired t-test, a sign test or a Wilcoxon sign-rank test with $\alpha= 1\%$. This result may surprise the reader, given how rarely rotation forest is used in experimental comparisons. However, we do not wish to oversell this result. The mean difference in accuracy between the tuned rotation forest and SVM is just 1.2\%, and the median difference is just over 0.5\%. Our main point is that rotation forest is a classifier that is clearly competitive with the most popular state-of-the-art classifiers, but has been largely ignored due to the poor choice of default parameters in the standard implementation (for example, Rotation Forest is not included as a Learner in the AutoWEKA package~\cite{kotthoff16autoweka}). We think this result merits further investigation, and a slight divergence from the main point of this paper, which is to try and stop people evaluating classifiers with default parameters. Rotation forest  has some fundamental differences to random forest that may not be well appreciated by the machine learning research community. Hence, we summarise how it works in Algorithm~\ref{algo:RotF}. In summary, for each tree, random forest partitions the feature set, performs a restricted PCA on each of these subsets (via class and case sampling), then recombines the features over the whole train set. One difference to random forest is that it does not use bagging, and hence we cannot utilise out of bag error for model selection. Another is that it does not do random feature selection, but rather uses all features for all trees.

\begin{algorithm}[!ht]
	\caption{buildRotationForest(Data $D$)}
\label{algo:RotF}
	\begin{algorithmic}[1]
\REQUIRE $r$, the number of trees, $f$, the number of features, $p$, the sample proportion
\STATE Let ${\bf F}=<F_1 \ldots F_r>$ be the C4.5 trees in the forest.
\FOR{$i \leftarrow  1$ to $r$}
\STATE Randomly partition the original features into $k$ subsets, each with $f$ features, denoted $<S_1 \ldots S_k>$.
\STATE Let $D_i$ be the train set for tree $i$, initialised to the original data, $D_i=D$.
\FOR {$j \leftarrow 1$ to $k$}
\STATE Select a non-empty subset of classes and extract only cases with those class. Each class has 0.5 probability of inclusion.
\STATE Draw a proportion $p$ of cases (without replacement) of those with the selected class value
\STATE Perform a Principal Component Analysis (PCA) on this subset of data
\STATE Apply the PCA transform built on this subset to the whole train set
\STATE Replace the features $S_j$ in $D_i$ with the PCA features.
\ENDFOR
\STATE Build C4.5 Classifier $F_i$ on transformed data $D_i$.
\ENDFOR
\end{algorithmic}
\end{algorithm}

\begin{figure}[!ht]
	\centering
\begin{tabular}{cc}
       \includegraphics[width=6cm, trim={4cm 10.5cm 0cm 0cm}]{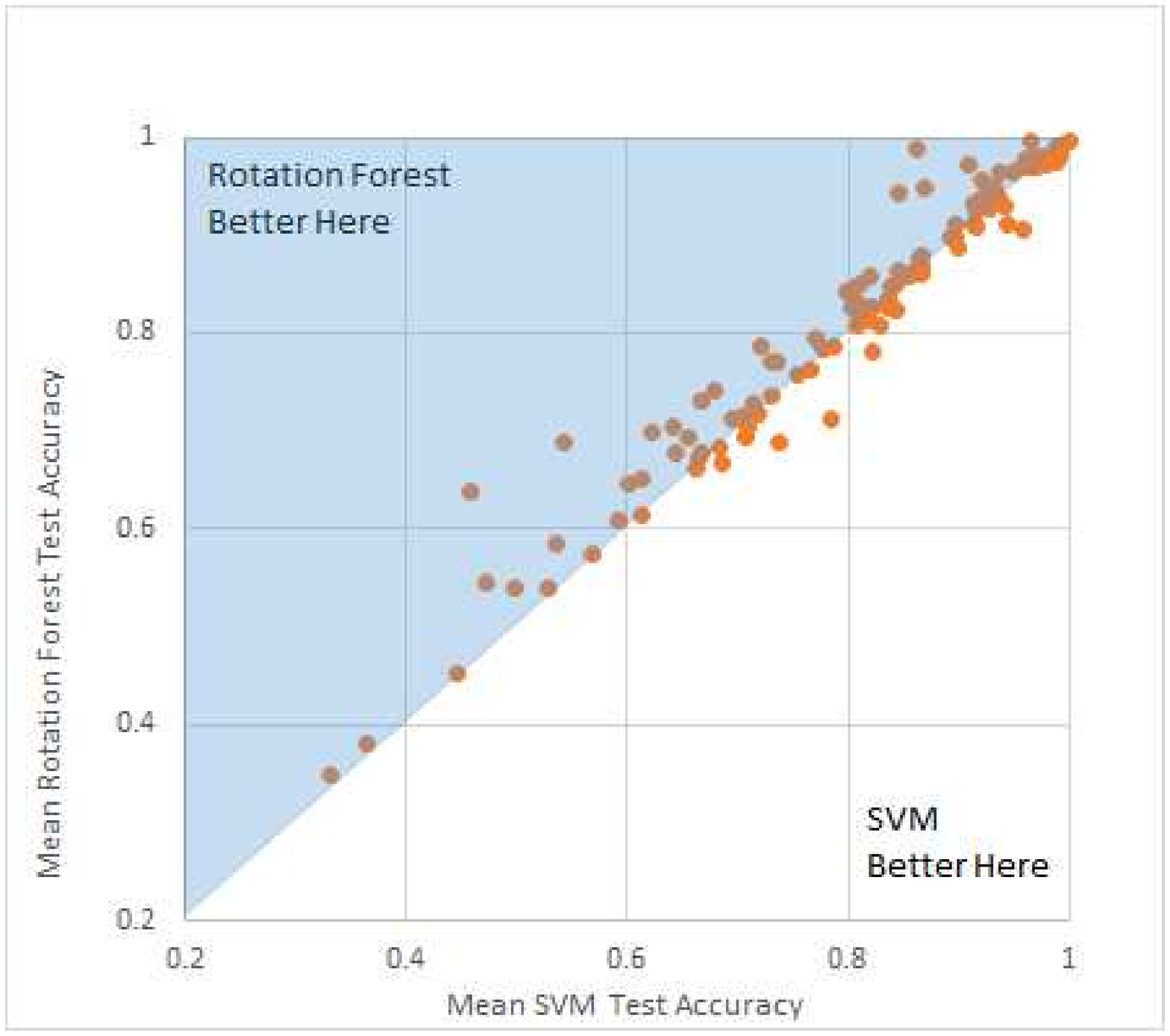}
       &
       \includegraphics[width=6cm,trim={4cm 10.5cm 0cm 0cm}]{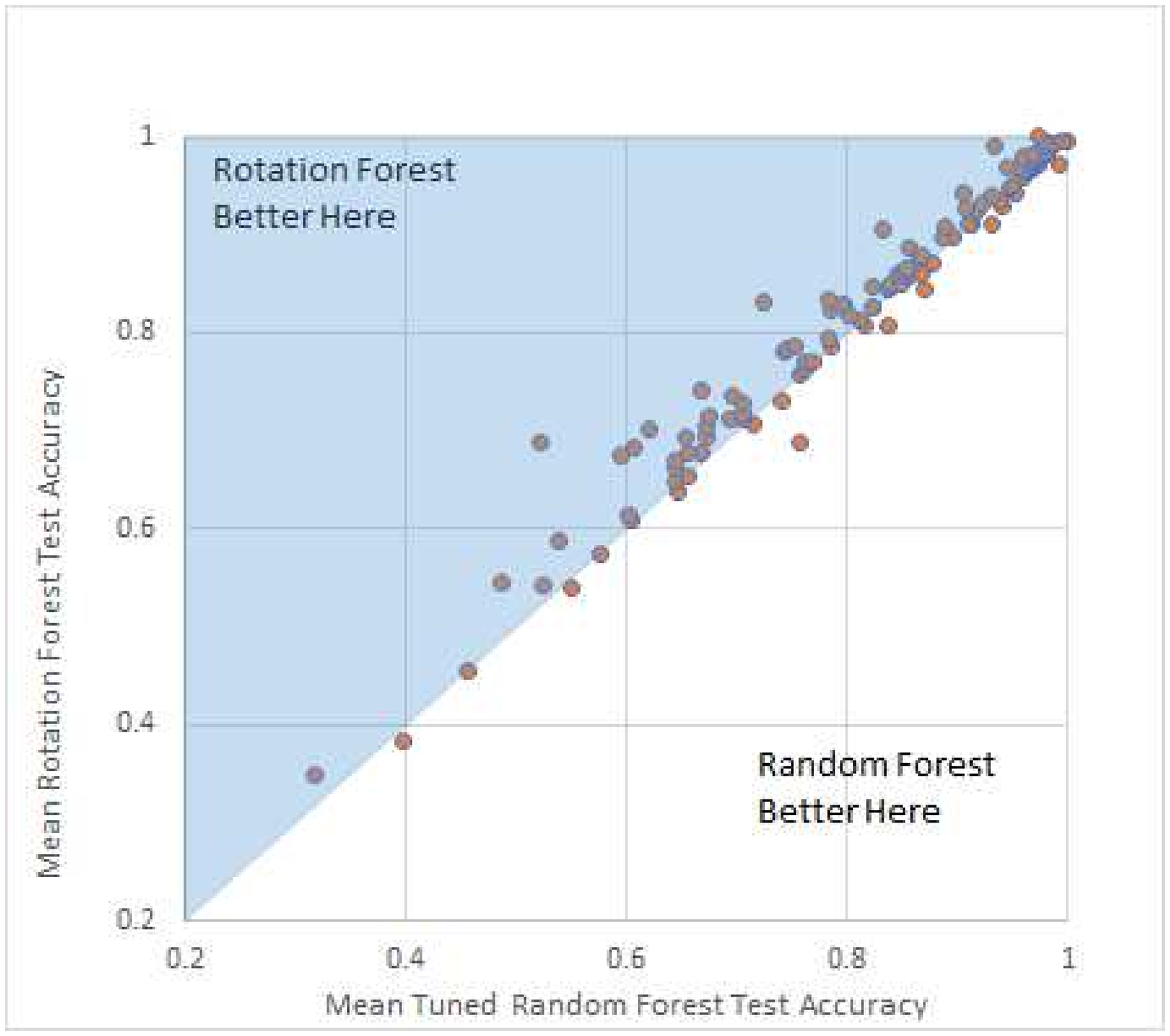}\\
       (a) & (b) \\
\end{tabular}
       	\caption{A scatter plot of accuracies of WEKA's SMO classifier with a RBF kernel and a linear kernel with default parameters. Figure (a) compares untuned classifiers, figure (b) plots the accuracy of tuned classifiers.}
       \label{fig:rotf}
\end{figure}

Figure~\ref{fig:rotf} shows the plots of rotation forest against SVM and random forest. Rotation forest beats SVM on 75 problems and loses on 35 (the remainder are ties). Rotation forest beats random forest on 87 problems and loses on 26.

Each problem involves 30 folds, hence it is possible to measure the variance across folds and perform two sample tests. We do not have the space to tabulate all the results\footnote{see bakeoff.xls for all results}. Instead we restrict Table~\ref{tab:bakeoff} to the problem where one of the three classifiers is significantly better (using a t-test) than the other two.

\begin{table}
\caption{Mean (and standard error) accuracy for three classifiers on problems where one classifier is significantly more accurate than the other two.}
\label{tab:bakeoff}
\begin{tabular}{|l|c|c|c|}\hline
Problem & Rotation Forest & SVM & Random Forest \\ \hline
arrhythmia	& {\bf 74.06 ($\pm$0.3)} & 	68.23 ($\pm$0.33)  & 	67 ($\pm$0.27) \\
audiology-std	&  {\bf 78.73 ($\pm$0.68)} & 	72.9 ($\pm$0.68)  & 	75.4 ($\pm$0.62) \\
balance-scale	& 90.54 ($\pm$0.15) & 	 {\bf 96.4 ($\pm$0.18)}  & 	83.34 ($\pm$0.3) \\
blood	&  {\bf 78.8 ($\pm$0.22)} & 	77.74 ($\pm$0.19)  & 	74.88 ($\pm$0.28) \\
breast-cancer-wisc	& {\bf 97.06 ($\pm$0.12)} & 	96.49 ($\pm$0.13)  & 	96.53 ($\pm$0.11) \\
conn-bench-vowel-detrending	& 97.71 ($\pm$0.16) & 	{\bf 98.98 ($\pm$0.1)}  & 	96.45 ($\pm$0.21) \\
contrac	& {\bf 54.17 ($\pm$0.23)} & 	52.64 ($\pm$0.3)  & 	52.47 ($\pm$0.25) \\
glass	& 73.06 ($\pm$0.53) & 	67.52 ($\pm$0.53)  & 	{\bf 75.69 ($\pm$0.57)} \\
hayes-roth	& 68.72 ($\pm$0.71) & 	74.32 ($\pm$0.62)  & {\bf 	77.33 ($\pm$0.65)} \\
hepatitis	& 80.94 ($\pm$0.39) & 	80.51 ($\pm$0.5)  & 	{\bf 82.99 ($\pm$0.51)}\\
hill-valley	& {\bf 68.76 ($\pm$0.34)} & 	53.15 ($\pm$0.41)  & 	53.15 ($\pm$0.37) \\
ilpd-indian-liver	& {\bf 71.27 ($\pm$0.11)} & 	69.83 ($\pm$0.35)  & 	69.54 ($\pm$0.35) \\
letter	& 96.05 ($\pm$0.04) & 	{\bf 96.63 ($\pm$0.04)}  & 	95.21 ($\pm$0.05) \\
molec-biol-promoter	& 84.38 ($\pm$0.82) & 	80.68 ($\pm$0.75)  & 	{\bf 87.47 ($\pm$0.58)} \\
molec-biol-splice	& 94.05 ($\pm$0.14) & 	84.18 ($\pm$0.16)  & 	{\bf 94.94 ($\pm$0.1)} \\
monks-1	& {\bf 98.36 ($\pm$0.62)} & 	87.06 ($\pm$0.58)  & 	92.66 ($\pm$0.45) \\
monks-2	& 70.41 ($\pm$0.75) & 	{\bf 78.58 ($\pm$0.58)}  & 	69.16 ($\pm$0.51) \\
musk-2	& 97.45 ($\pm$0.07) & 	{\bf 98.97 ($\pm$0.05)}  & 	97.14 ($\pm$0.06) \\
nursery	& {\bf 99.57 ($\pm$0.03)} & 	99.22 ($\pm$0.03)  & 	99.09 ($\pm$0.03) \\
optical	& 98.18 ($\pm$0.03) & 	{\bf 99.05 ($\pm$0.03)}  & 	97.99 ($\pm$0.04) \\
pendigits	& 99.23 ($\pm$0.02) & {\bf 	99.49 ($\pm$0.02)}  & 	98.81 ($\pm$0.03) \\
pittsburg-bridges-TYPE	& {\bf 	57.65 ($\pm$0.71)} & 	52.59 ($\pm$1.07)  & 	52.84 ($\pm$1.08) \\
ringnorm	& 97.77 ($\pm$0.04) & 	{\bf 98.58 ($\pm$0.03)}  & 	96.07 ($\pm$0.06) \\
semeion	& 90.93 ($\pm$0.21) & 	{\bf 94.38 ($\pm$0.16)}  & 	92.52 ($\pm$0.15) \\
soybean	& {\bf 92.94 ($\pm$0.23)} & 	91.27 ($\pm$0.26)  & 	91.59 ($\pm$0.26) \\
statlog-australian-credit	& {\bf 67.72 ($\pm$0.17)} & 	66.85 ($\pm$0.22)  & 	66.52 ($\pm$0.29) \\
statlog-vehicle	& 78.33 ($\pm$0.29) & 	{\bf 82.05 ($\pm$0.3)}  & 	74.7 ($\pm$0.25) \\
thyroid	& {\bf 99.48 ($\pm$0.02)} & 	96.59 ($\pm$0.05)  & 	99.4 ($\pm$0.02) \\
tic-tac-toe	& 98.04 ($\pm$0.18) & 	{\bf 99.29 ($\pm$0.1)}  & 	97.72 ($\pm$0.18) \\
wall-following	& 96.95 ($\pm$0.07) & 	90.8 ($\pm$0.12)  & 	{\bf 99.2 ($\pm$0.03)} \\
waveform	& 86.13 ($\pm$0.09) & 	{\bf 86.63 ($\pm$0.09)}  & 	84.91 ($\pm$0.12) \\
waveform-noise	& {\bf 86.55 ($\pm$0.08)} & 	86.09 ($\pm$0.08)  & 	85.25 ($\pm$0.09) \\
yeast	& {\bf 61.07 ($\pm$0.28)} & 	59.18 ($\pm$0.24)  & 	59.8 ($\pm$0.31) \\
\hline
 \end{tabular}
\end{table}

Table~\ref{tab:bakeoff} shows that rotation forest beats the other two on 15 problems, SVM wins on 12 and random forest on 6. On some of the problems (e.g. nursery, letter and breast-cancer-wisc), the difference is very small, albeit significant. However, on other problems, there is a really large difference. For example, rotation forest is 15\% more accurate than both SVM and random forest on hill-valley, whereas SVM has an 8.5\% advantage on monks-2. This leads us to ask whether we could accurately choose between classifiers based on the train data alone.

\subsection{Can We Choose Which Algorithm to Use Via Cross-Validation on the Training Set?}

\begin{figure}[!ht]
	\centering
       \includegraphics[width=10cm, trim={0cm 9.5cm 0cm 2cm}]{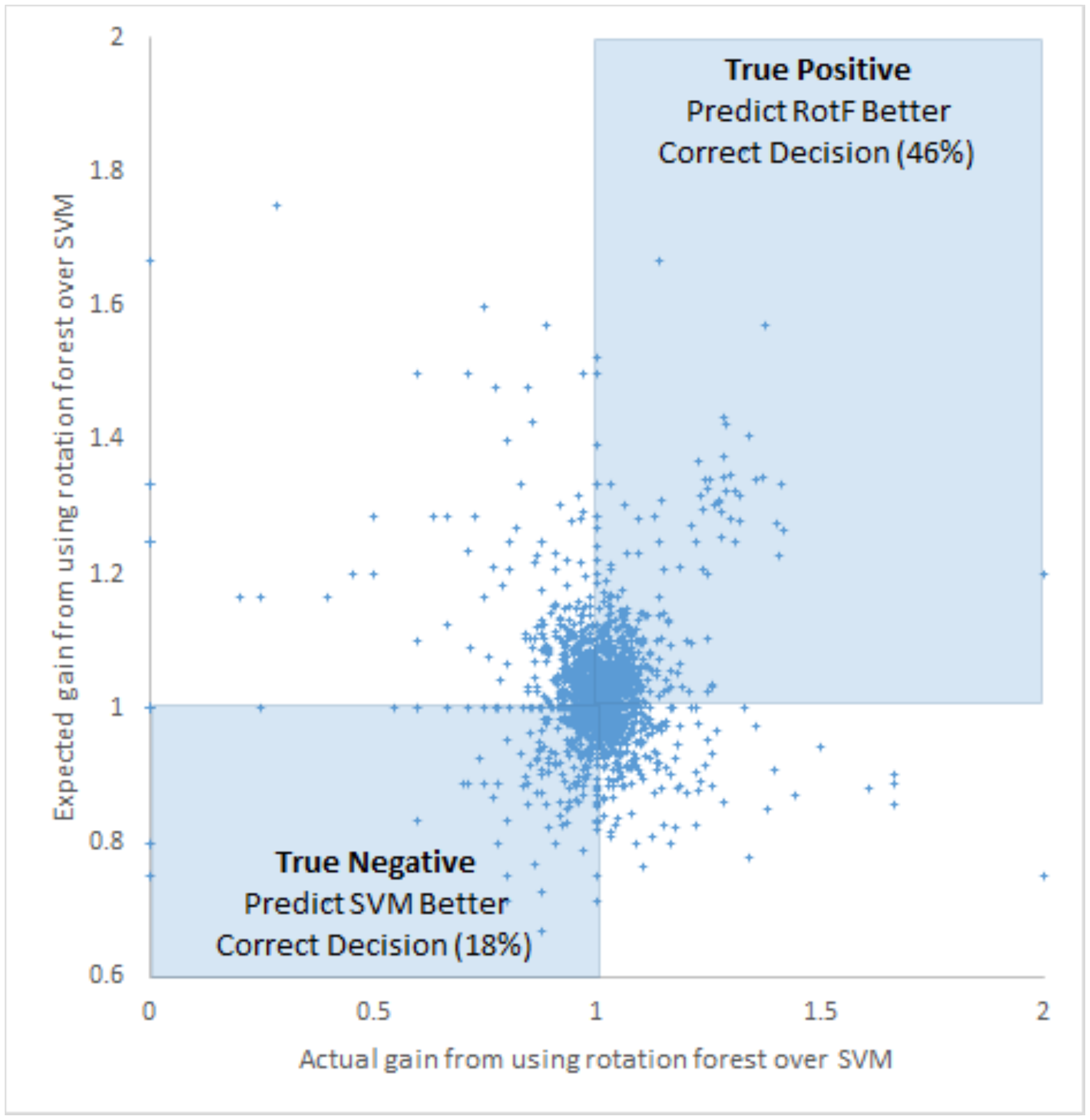}
       	\caption{Texas sharpshooter plot for rotation forest against SVM. Each point represents a single fold on a single problem. The y-axis is the ratio of cross-validated train accuracy, the x-axis the ratio of test accuracy. }
       \label{fig@texas}
\end{figure}

Knowing which algorithm is more accurate on average over multiple data is of interest, as it gives a reasonable default position. Based on our experiments, we would recommend using a rotation forest with the number of trees set through cross validation as a default classifier. However, practitioners are ultimately interested in deciding which algorithm to use for a specific problem. The global differences between the three algorithms we evaluate is small and the range of differences is large, hence our default position may not be that useful for specific classification tasks. Suppose we took the decision to choose rotation forest over SVM on every resample of every data set. We would have only made the correct decision 52\% of the time (ignoring ties). Could we can make a better choice of classifier based on train set accuracy? We address the question using a technique first proposed in~\cite{batista14cid}.
The basic principle is that the ratio of the cross-validation accuracy (over the training set) of two classifiers should give an indication to the outcome for the test data. However, if the cross validation accuracy is biased (for instance by use in tuning the parameters) or subject to high variance (due to the use of an overly complex model structure), then often the ratio will be misleading. The plot of cross-validation accuracy ratio vs. testing accuracy ratio gives a continuous form of contingency table for assessing the usefulness of the training accuracy. If the ratio on cross-validation and testing data are both greater than one then the case is true positive (we predict a gain for Classifier A and also observe a gain). If both ratios are less than one, the problem is a true negative (we predict a loss for Classifier A and also observe a loss). Otherwise, we have an undesirable outcome. If the data sets are evenly spread between the four quadrants, then Batista {\em et al.}~\cite{batista14cid} observe that we have a situation analogous to the Texas sharpshooter fallacy (which comes from a joke about a Texan who fires shots at the side of a barn, then paints a target centred on the biggest cluster of hits and claims to be a sharpshooter). Figure~\ref{fig@texas} shows the Texas sharpshooter plot for rotation forest against SVM.

We have improved our decision making, but not decisively. We now make the correct decision 64\% of the time, compared to our default position of 52\%. This highlights the danger of relying too much on train set accuracy, particularly when it has been optimised on the train data through a parameter search. We believe   there is scope for research into better mechanisms for choosing a classifier.

\section{Can We Set Better Default Parameters?}
\label{better_defaults}

A parameter search of some kind clearly improves the three classifiers we are evaluating, but such a search is not always feasible due to the size of the training data and limited computing resources. This raises the question of whether we can find better global default values than those currently used, especially in WEKA, and whether setting default parameters based on the characteristics of the data can lead to a better classifier.

\begin{figure}[!ht]
	\centering
       \includegraphics[width=10cm, trim={2cm 10cm 2cm 9cm}]{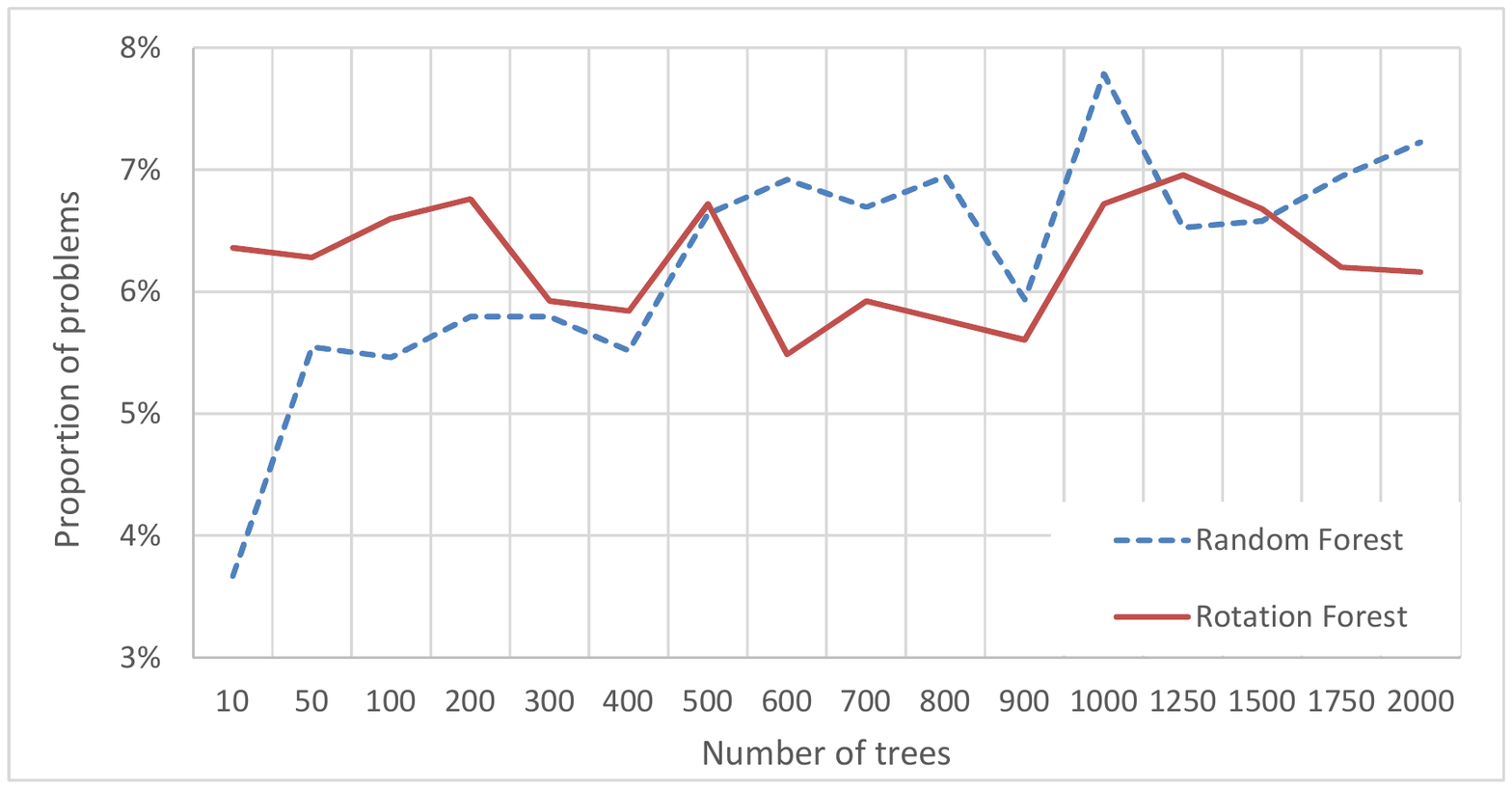}
       	\caption{Proportion of resamples (over all folds) where different number of trees gave the optimal train accuracy.}
       \label{fig:numTrees}
\end{figure}

Figure~\ref{fig:numTrees} shows the proportion of resamples in which a particular tree size was selected over all problems for both random forest and rotation forest. There seems to be very little pattern, although the standard default of 500 trees for Random Forest seems reasonable. Similar wide variation was observed within samples from the same problem. However, it is worth noting that there may be very little difference in accuracy between classifiers with different number of trees and that if more than one setting had identical training accuracy, we chose randomly. This happens frequently, and we believe that both forest techniques are more robust to this parameter and SVM is to the RBF parameters.

\begin{figure}[!ht]
	\centering
       \includegraphics[width=10cm, trim={2cm 3cm 2cm 4cm}]{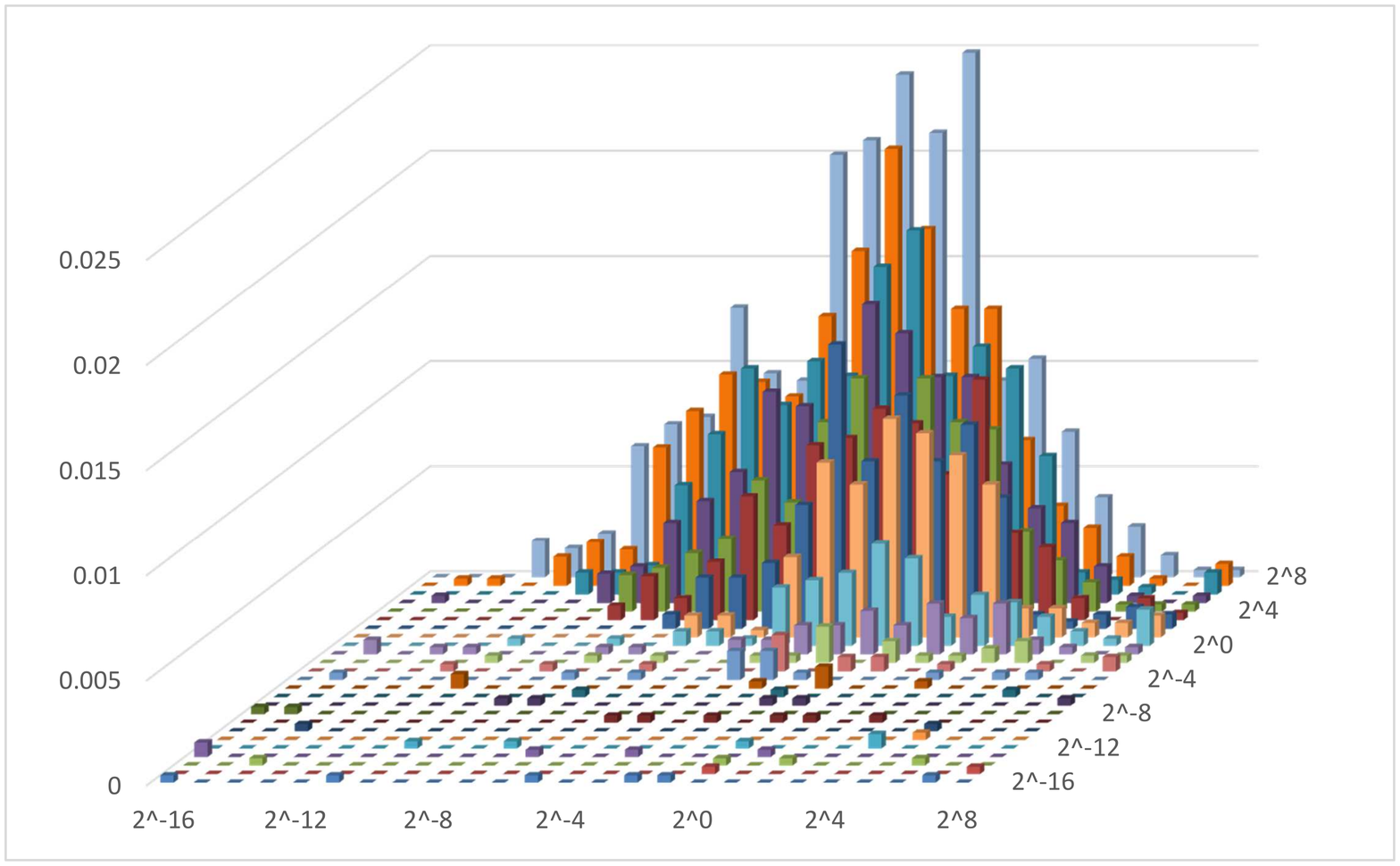}
       	\caption{Proportion of resamples (over all folds) where different combinations of parameters $C$ and $\gamma$ gave the optimal train accuracy.}
       \label{fig:svmParas}
\end{figure}

Figure~\ref{fig:svmParas} shows the distribution of parameter combinations selected for the tuned SVM. There is much  tighter grouping around values of $C$ at the higher end of the range and $\gamma$ in the middle. This indicates that SVM is much more sensitive to parameter values than the forest algorithms, and again emphasises the need for tuning. It also suggests that a large value of $C$, such as 256, and a middle range value for $\gamma$, such as $0.25$ are much better defaults for the RBF kernel than the current WEKA defaults of $C=1$ and $\gamma=0.01$.

\section{Conclusion}
\label{conclusions}

Using a weak straw man is one of the easiest ways of inflating the significance of the gain offered by a new algorithm. Comparison against a classifier such as logistic regression is likely to draw reviewer criticism, whereas comparison against some form of support vector machine or random forest will often be accepted as sufficient, particularly given recent experimental support for these algorithms~\cite{delgado14hundreds}. However, untuned SVM and random forest can be no better than basic classifiers, and as such do not constitute a reasonable representation of state-of-the-art in terms of classification accuracy.

Our basic point is that if you compare a new classifier against one with default parameters, you can have no real confidence in any improvement you detect, even if it is seemingly significant. Through this evaluation, we have also highlighted how a highly competitive classifier such as rotation forest is often ignored because the default parameters almost universally used are worse than any sensible possible setting. We are not claiming rotation forest is a better classifier than SVM or random forest. We do claim that it should be considered as state-of-the-art and that on average, over a large number of problems, it is significantly more accurate than the variants of SVM and random forest we have evaluated.

Accuracy is not everything when it comes to evaluating a classification algorithm. An algorithm does not have to be significantly better than state-of-the-art on average over all problems in a test suite to be of interest. It may work better on data with certain characteristics, such as a large number of attributes, or data from certain problem domains. However, we maintain that it is up to those proposing algorithms to identify the scenario under which they add value. Classification research is a mature area; there are three families of algorithms that dominate: support vector machines and related kernel methods; tree based ensemble techniques; and multi-layer perceptron/deep learning algorithms. If researchers add to the huge quantity of research into these algorithmic domains without a thorough experimental evaluation they risk just adding to the background noise. A clever idea is not enough to interest practitioners who already have an extensive armoury of algorithms to employ for new classification problems.

An obvious gap in this work is the lack of consideration of any form of deep learning style of neural network (NN). NN are notoriously sensitive to parameter settings; An article in the New York Times in relation to Google's translation deep learning approach~\footnote{http://www.nytimes.com/2016/12/14/magazine/the-great-ai-awakening.html?\_r=0}, states that ``So much of what they [the Google team] did was just gut. How many neurons per layer did you use? 1,024 or 512? How many layers? How many sentences did you run through at a time? How long did you train for?". Mike Schuster is quoted as saying ``You're always saying: When do we stop? How do I know I'm done? You never know you’re done .... It's a little bit an art — where you put your brush to make it nice. It comes from just doing it. Some people are better, some worse.'' It would be highly desirable to both assess the importance of these parameters on standard classification problems and attempt to automate this process, as not all users will be equally skilful.

To researchers working on algorithm development, we suggest that there are several areas of research that relate to rotation forests that could yield improvements. We suggest three areas of investigation. Firstly, model selection could be made much faster. The number of trees for random forest can be selected at almost no overhead by using the out-of-bag error and incrementally adding trees. It would be beneficial if rotation forest could be adapted to do the same without significant loss of accuracy. Secondly, rotation forest could be adapted to use feature selection. Rotation forest uses all features for every tree, but for large feature spaces this may introduce excessive time overhead and may decrease accuracy. If a feature selection scheme could be incorporated without loss of accuracy, that would help advance the field. Finally, we believe there is scope for adapting rotation forest to better work with discrete data; PCA makes little sense for categorical variables, the data set we have used have all been converted to be all real valued. We believe performance could be improved on categorical data by employing alternative filters.

To practitioners, if resources permit, we recommend using tuned versions of all three classifiers, then selecting a classifier based on cross validated training accuracy. Be warned, however, that this measure is not completely reliable. If this is infeasible, a rotation forest of 500 or 1000 trees is likely to perform best on average. If even this is too slow or memory intensive, then a random forest with the standard defaults of 500 trees and $\sqrt{m}$ features for each tree is probably going to give the most reliable results.

\section*{Acknowledgment}
This work is supported by the UK Engineering and Physical Sciences Research Council (EPSRC)  [grant number EP/ M015087/1]. The experiments were carried out on the High Performance Computing Cluster supported by the Research and Specialist Computing Support service at the University of East Anglia.

\appendix

\section{Methodologies Adopted by Muppet Labs}
\label{sec:muppet}

Table~\ref{tbl:datasets} provides information on the seventeen benchmark datasets used by Dr  Honeydew and Beaker.  These are largely taken from the suite used by R\"atsch \emph{et al.}  \cite{Raetsch2001a}, augmented by Ripley's \texttt{synthetic} benchmark dataset  \cite{Ripley1996a} and the \texttt{ionosphere}, \texttt{sonar} and \texttt{vertebra} datasets  from the UCI repository \cite{Bay1999a}.  For each dataset there are 100 random partitions of  the data to form training and test sets (20 in the case of the larger \texttt{image} and \texttt{splice} benchmarks).

\begin{table}[tbh!]
   \caption{Details of datasets used in the Muppet experiment.}
   \newlength\colwidthi
   \newlength\colwidthii
   \newlength\colwidthiii
   \newlength\colwidthiv
   \newlength\colwidthv
   \settowidth\colwidthi{\bfseries Breast can}
   \settowidth\colwidthii{1300}
   \settowidth\colwidthiii{7000}
   \settowidth\colwidthiv{100}
   \settowidth\colwidthv{60}
   \newcommand{\colentry}[2]{%
   \makebox[\csname colwidth\romannumeral#1\endcsname][r]{#2}}
   \renewcommand{\arraystretch}{1.2}
   \begin{center}
      \begin{tabular}{|l|c|c|c|c|}
         \hline
         \multirow{2}{\colwidthi}{\hfil\bfseries Dataset} &
         \bfseries Training &
         \bfseries Testing &
         \bfseries Number of &
         \bfseries Input \\
          &
         \bfseries Patterns &
         \bfseries Patterns &
         \bfseries Replicates &
         \bfseries Features \\
         \hline
         \hline
         \textbf{Banana}        & \colentry2{400}  & \colentry3{4900} & \colentry4{100} & \colentry5{2} \\
         \textbf{Breast cancer} & \colentry2{200}  & \colentry3{77}   & \colentry4{100} & \colentry5{9} \\
         \textbf{Diabetis}      & \colentry2{468}  & \colentry3{300}  & \colentry4{100} & \colentry5{8} \\
         \textbf{Flare solar}   & \colentry2{666}  &  \colentry3{400} & \colentry4{100} &  \colentry5{9} \\
         \textbf{German}        &  \colentry2{700} &  \colentry3{300} & \colentry4{100} & \colentry5{20} \\
         \textbf{Heart}         &  \colentry2{170} &  \colentry3{100} & \colentry4{100} & \colentry5{13} \\
         \textbf{Image}         & \colentry2{1300} & \colentry3{1010} &  \colentry4{20} & \colentry5{18} \\
         \textbf{Ionosphere}    &  \colentry2{200} & \colentry3{151}  & \colentry4{100} & \colentry5{34} \\
         \textbf{Ringnorm}      &  \colentry2{400} & \colentry3{7000} & \colentry4{100} & \colentry5{20} \\
         \textbf{Sonar}         & \colentry2{138} & \colentry3{70}    & \colentry4{100} & \colentry5{60} \\
         \textbf{Splice}        & \colentry2{1000} & \colentry3{2175} &  \colentry4{20} & \colentry5{60} \\
         \textbf{Synthetic}     &  \colentry2{250} & \colentry3{1000} & \colentry4{100} & \colentry5{2} \\
         \textbf{Thyroid}       &  \colentry2{140} &   \colentry3{75} & \colentry4{100} &  \colentry5{5} \\
         \textbf{Titanic}       &  \colentry2{150} & \colentry3{2051} & \colentry4{100} &  \colentry5{3} \\
         \textbf{Twonorm}       &  \colentry2{400} & \colentry3{7000} & \colentry4{100} & \colentry5{20} \\
         \textbf{Vertebra}      &  \colentry2{248} & \colentry3{62}   & \colentry4{100} & \colentry5{6} \\
         \textbf{Waveform}      &  \colentry2{400} & \colentry3{4600} & \colentry4{100} & \colentry5{21} \\
         \hline
      \end{tabular}
   \end{center}
   \label{tbl:datasets}
\end{table}

Dr Bunsen Honeydew used the LS-SVM with RBF kernel (implemented using the Generalised Kernel Machine \cite{Cawley2007b} toolbox\footnote{\url{http://theoval.cmp.uea.ac.uk/projects/gkm/}} for MATLAB; regularisation parameter $\gamma = 1$, kernel parameter $\eta = 1$), the SVM with RBF kernel (implemented using a MATLAB SVM toolbox\footnote{\url{http://theoval.cmp.uea.ac.uk/svm/toolbox/}} \cite{Cawley2000a}; regularisation parameter $C = 1$, kernel parameter $\eta = 1$), the EP-GPC (implemented by the Gaussian Processes for Machine Learning MATLAB toolbox\footnote{http://www.gaussianprocess.org/gpml/code/matlab/doc/}; mean function $=$ \texttt{meanZero}, likelihood function $=$ \texttt{likErf}, covariance function $=$ \texttt{covSEiso}, (log) covariance function hyper-parameters $= [0, 0]$).  The [ML]$^2$A is, in fact, a simple multi-layer perceptron neural network with Bayesian regularisation \cite{Bishop1995a} (implemented using the NETLAB\footnote{http://www.aston.ac.uk/eas/research/groups/ncrg/resources/netlab/} toolbox \cite{Nabney2004b} for MATLAB; to guard against problems associated with local minima, eight MLPs are trained and the network with the highest marginal likelihood used to make predictions);

Beaker used the same toolboxes, models and experimental protocol as Dr Honeydew, however in each case, the hyper-parameters were tuned, starting from the default values used by Dr Honeydew, using a simple grid search procedure.  For the LS-SVM, the grid search was performed over values of $\log_2\gamma$ and $\log_2\eta$ from -20 to +12 in increments of 1, minimising the virtual leave-one-out cross-validation estimate of the mean squared error \cite{Cawley2006a}.  For the SVM, the grid over $\log_2C$ and $\log_2\eta$ extended from -16 to 16 in increments of 1, minimising a ten-fold cross-validation estimate of the error rate.  For the EP-GPC, the hyper-parameter grid spanned the range -4 to +10 in increments of 0.5, maximising the marginal likelihood of the model.


\begin{thebibliography}{10}

\bibitem{oshiro12nowmany}
T.~Oshiro \and P.~Perez and J.~Baranauskas.
\newblock {\em How Many Trees in a Random Forest?}, pages 154--168.
\newblock 2012.

\bibitem{bagnall16bakeoff}
A.~Bagnall, J.~Lines, A.~Bostrom, J.~Large, and E.~Keogh.
\newblock The great time series classification bake off: a review and
  experimental evaluation of recent algorithmic advances.
\newblock {\em Data Mining and Knowledge Discovery}, Online first, open access,
  2016.

\bibitem{bagnall15cote}
A.~Bagnall, J.~Lines, J.~Hills, and A.~Bostrom.
\newblock Time-series classification with cote: The collective of
  transformation-based ensembles.
\newblock {\em {IEEE} Transactions on Knowledge and Data Engineering},
  27:2522--2535, 2015.

\bibitem{batista14cid}
G.~Batista, E.~Keogh, O.~Tataw, and V.~deSouza.
\newblock {CID}: an efficient complexity-invariant distance measure for time.
\newblock {\em Data Mining and Knowledge Discovery}, 28(3):624--669, 2014.

\bibitem{Bay1999a}
S.~D. Bay.
\newblock The {UCI} {KDD} archive $[$\texttt{http://kdd.ics.uci.edu/}$]$.
\newblock University of California, Department of Information and Computer
  Science, Irvine, CA, 1999.

\bibitem{benavoli16posthoc}
A.~Benavoli, G.~Corani, and F.~Mangili.
\newblock Should we really use post-hoc tests based on mean-ranks?
\newblock {\em Journal of Machine Learning Research}, 17:1--10, 2016.

\bibitem{Bishop1995a}
C.~M. Bishop.
\newblock {\em Neural Networks for Pattern Recognition}.
\newblock Oxford University Press, 1995.

\bibitem{Boser1992a}
B.~E. Boser, I.~M. Guyon, and V.~Vapnik.
\newblock A training algorithm for optimal margin classifiers.
\newblock In D.~Haussler, editor, {\em Proceedings of the fifth Annual ACM
  Workshop on Computational Learning Theory}, pages 144--152, Pittsburgh, PA,
  July 1992.

\bibitem{Cawley2000a}
G.~C. Cawley.
\newblock {MATLAB} support vector machine toolbox (v0.55$\beta$) $[$
  \texttt{http://theoval.sys.uea.ac.uk/\~{}gcc/svm/toolbox}$]$.
\newblock University of East Anglia, School of Information Systems, Norwich,
  Norfolk, U.K. NR4 7TJ, 2000.

\bibitem{Cawley2006a}
G.~C. Cawley.
\newblock Leave-one-out cross-validation based model selection criteria for
  weighted {LS-SVM}s.
\newblock In {\em Proceedings of the {IEEE/INNS} International Joint Conference
  on Neural Networks ({IJCNN-06})}, pages 1661--1668, Vancouver, BC, Canada,
  July~16--21 2006.

\bibitem{Cawley2007b}
G.~C. Cawley, G.~J. Janacek, and N.~L.~C. Talbot.
\newblock Generalised kernel machines.
\newblock In {\em Proceedings of the {IEEE/INNS} International Joint Conference
  on Neural Networks ({IJCNN-07})}, pages 1720--1725, Orlando, Florida, USA,
  August~12--17 2007.

\bibitem{Cawley2010b}
G.~C. Cawley and N.~L.~C. Talbot.
\newblock Over-fitting in model selection and subsequent selection bias in
  performance evaluation.
\newblock {\em Journal of Machine Learning Research}, 11:2079--2107, July 2010.

\bibitem{Cortes1995a}
C.~Cortes and V.~Vapnik.
\newblock Support-vector networks.
\newblock {\em Machine Learning}, 20(3):273--297, September 1995.

\bibitem{cox58logistic}
D.~Cox.
\newblock The regression analysis of binary sequences (with discussion).
\newblock {\em Journal of the Royal Statistical Society, B}, 20:215--242, 1958.

\bibitem{Demsar2006a}
J.~Dem\v{s}ar.
\newblock Statistical comparisons of classifiers over multiple data sets.
\newblock {\em Journal of Machine Learning Research}, 7:1--30, 2006.

\bibitem{delgado14hundreds}
M.~Fern\'{a}ndez-Delgado, E.~Cernadas, S.~Barro, and D.~Amorim.
\newblock Do we need hundreds of classifiers to solve real world classification
  problems?
\newblock {\em Journal of Machine Learning Research}, 15:3133--3181, 2014.

\bibitem{garcia08pairwise}
S.~Garc\'{i}a and F.~Herrera.
\newblock An extension on “statistical comparisons of classifiers over
  multiple data sets” for all pairwise comparisons.
\newblock {\em Journal of Machine Learning Research}, 9:2677--2694, 2008.

\bibitem{kuhn08caret}
M.~Hall, E.~Frank, G.~Holmes, B.~Pfahringer, P.~Reutemann, and I.H. Witten.
\newblock Building predictive models in {R} using the caret package.
\newblock {\em ACM SIGKDD Explorations Newsletter}, 11(1):10--18, 2009.

\bibitem{hall2009weka}
M.~Hall, E.~Frank, G.~Holmes, B.~Pfahringer, P.~Reutemann, and I.H. Witten.
\newblock The {WEKA} data mining software: an update.
\newblock {\em ACM SIGKDD Explorations Newsletter}, 11(1):10--18, 2009.

\bibitem{Hand2006a}
D.~J. Hand.
\newblock Classifier technology and the illusion of progress.
\newblock {\em Statistical Science}, 21(1):1--14, 2006.

\bibitem{keerthi2003asymptotic}
S~Sathiya Keerthi and Chih-Jen Lin.
\newblock Asymptotic behaviors of support vector machines with gaussian kernel.
\newblock {\em Neural computation}, 15(7):1667--1689, 2003.

\bibitem{kohavi95wrappers}
R.~Kohavi.
\newblock {\em Wrappers for Performance Enhancement and Oblivious Decision
  Graphs}.
\newblock PhD thesis, Stanford University, Department of Computer Science,
  Stanford University, 1995.

\bibitem{kotthoff16autoweka}
L.~Kotthoff, C.~Thornton, H.~Hoos, F.~Hutter, and K.~Leyton-Brown.
\newblock {Auto-WEKA 2.0}: Automatic model selection and hyperparameter
  optimization in {WEKA}.
\newblock {\em Journal of Machine Learning Research}, 17:1--5, 2016.

\bibitem{lines16hivecote}
J.~Lines, S.~Taylor, and A.~Bagnall.
\newblock Hive-cote: The hierarchical vote collective of transformation-based
  ensembles for time series classification.
\newblock In {\em Proceedings of the {IEEE} International Conference on Data
  Mining}, 2016.

\bibitem{Nabney2004b}
I~Nabney.
\newblock {\em {NETLAB}: Algorithms for pattern recognition}.
\newblock Advances in Pattern Recognition. Springer, 2004.

\bibitem{platt98sequential}
J.~Platt.
\newblock Fast training of support vector machines using sequential minimal
  optimization.
\newblock In B.~Schoelkopf, C.~Burges, and A.~Smola, editors, {\em Advances in
  Kernel Methods - Support Vector Learning}. 1998.

\bibitem{Rasmussen2006a}
C.~E. Rasmussen and C.~K.~I. Williams.
\newblock {\em {G}aussian Processes for Machine Learning}.
\newblock Adaptive Computation and Machine Learning. MIT Press, 2006.

\bibitem{Raetsch2001a}
G.~R\"atsch, T.~Onoda, and K.-R. M\"uller.
\newblock Soft margins for {A}da{B}oost.
\newblock {\em Machine Learning}, 42(3):287--320, March 2001.

\bibitem{Ripley1996a}
B.~D. Ripley.
\newblock {\em Pattern Recognition and Neural Networks}.
\newblock Cambridge University Press, 1996.

\bibitem{rodriguez06rotation}
J.J. Rodriguez, L.I. Kuncheva, and C.J. Alonso.
\newblock Rotation forest: A new classifier ensemble method.
\newblock {\em IEEE Trans. Pattern Analysis and Machine Intelligence},
  28(10):1619--1630, 2006.

\bibitem{Suykens2002a}
J.~A.~K. Suykens, T.~Van~Gestel, J.~De~Brabanter, B.~De~Moor, and
  J.~Vanderwalle.
\newblock {\em Least squares support vector machine}.
\newblock World Scientific Publishing Company, Singapore, 2002.

\bibitem{Wainberg16randomForests}
M.~Wainberg, B.~Alipanahi, and B.~Frey.
\newblock Are random forests truly the best classifiers?
\newblock {\em Journal of Machine Learning Research}, 17(110):1--5, 2016.

\end{thebibliography}
\end{document}